\documentclass[conference]{IEEEtran}   
\pdfoutput=1

\IEEEoverridecommandlockouts
\usepackage{cite}
\usepackage[ruled,vlined]{algorithm2e}

\usepackage{amsmath,amssymb,amsfonts}
\usepackage{graphicx}
\usepackage{textcomp}
\usepackage{xcolor}
\usepackage[bottom=0.653in,top=0.958in, right=0.653in, left=0.653in]{geometry}
\def\BibTeX{{\rm B\kern-.05em{\sc i\kern-.025em b}\kern-.08em
    T\kern-.1667em\lower.7ex\hbox{E}\kern-.125emX}}
\begin{document}


\title{Performance Improvement of Path Planning algorithms with Deep Learning Encoder Model\\
\thanks{This study was financed in part by the Coordena\c{c}\~ao de Aperfei\c{c}oamento de Pessoal de N\'ivel Superior - Brasil (CAPES) - Finance Code 001, and the Brazilian agencies FACEPE and CNPq.}
}

\author{\IEEEauthorblockN{1\textsuperscript{st} Janderson Ferreira}
\IEEEauthorblockA{\textit{dept. Computer Engineering.} \\
\textit{Polytechnic School (UPE)}\\
Recife, Brazil \\
jrb@ecomp.poli.br}
\and
\IEEEauthorblockN{2\textsuperscript{nd} Agostinho A. F. J\'unior}
\IEEEauthorblockA{\textit{dept. Computer Engineering.} \\
\textit{Polytechnic School (UPE)}\\
Recife, Brazil \\
aafj@ecomp.poli.br}
\and
\IEEEauthorblockN{3\textsuperscript{rd} Yves M. Galv\~ao}
\IEEEauthorblockA{\textit{dept. Computer Engineering.} \\
\textit{Polytechnic School (UPE)}\\
Recife, Brazil \\
ymg@ecomp.poli.br}
\and
\IEEEauthorblockN{4\textsuperscript{th} Pablo Barros}
\IEEEauthorblockA{\textit{Cognitive Architecture for Collaborative Technologies} \\
\textit{Istituto Italiano di Tecnologia (IIT)}\\
Genova, Italy \\
pablo.alvesdebarros@iit.it}
\and
\IEEEauthorblockN{5\textsuperscript{th} Sergio Murilo Maciel Fernandes}
\IEEEauthorblockA{\textit{dept. Computer Engineering.} \\
\textit{Polytechnic School (UPE)}\\
Recife, Brazil \\
smmf@ecomp.poli.br}
\and
\IEEEauthorblockN{6\textsuperscript{th} Bruno J. T. Fernandes}
\IEEEauthorblockA{\textit{dept. Computer Engineering.} \\
\textit{Polytechnic School (UPE)}\\
Recife, Brazil \\
bjtf@ecomp.poli.br}
}

\maketitle

\begin{abstract}
Currently, path planning algorithms are used in many daily tasks. They are relevant to find the best route in traffic and make autonomous robots able to navigate. The use of path planning presents some issues in large and dynamic environments. Large environments make these algorithms spend much time finding the shortest path. On the other hand, dynamic environments request a new execution of the algorithm each time a change occurs in the environment, and it increases the execution time. The dimensionality reduction appears as a solution to this problem, which in this context means removing useless paths present in those environments. Most of the algorithms that reduce dimensionality are limited to the linear correlation of the input data. Recently, a Convolutional Neural Network (CNN) Encoder was used to overcome this situation since it can use both linear and non-linear information to data reduction. This paper analyzes in-depth the performance to eliminate the useless paths using this CNN Encoder model. To measure the mentioned model efficiency, we combined it with different path planning algorithms. Next, the final algorithms (combined and not combined) are checked in a database that is composed of five scenarios. Each scenario contains fixed and dynamic obstacles. Their proposed model, the CNN Encoder, associated to other existent path planning algorithms in the literature, was able to obtain a time decrease to find the shortest path in comparison to all path planning algorithms analyzed. the average decreased time was 54.43\%.


\end{abstract}
1

\section{Introduction}
Path planning algorithms are essential for the accomplishment of many activities in different areas, for example, robot navigation \cite{kakoty2019mobile}, path apps for locomotion in cities (for pedestrian and driver) \cite{pang2019spath}, autonomous-driver cars \cite{zhu2018human}. These algorithms have different approaches to treat spatial information, the most used in the literature are,
Grid-based search (which transforms the environment in a grid-mesh) \cite{grid}, Interval-based search (similar to grid-based search it but uses space data instead of a grid) \cite{interval} and Reward-based (similar to a reinforcement learning in deep learning) \cite{reward}.

Based on Grid-based search, the first path planning algorithm was proposed by Dijkstra in 1956. Although this solution is always able to find the shortest path between two points, Dijkstra's algorithm has become obsolete because it has very high computational complexity. Considering the response time of the Dijkstra algorithm, it would be infeasible to be applied in many scenarios.

Given this problem, new algorithms with different approaches were created to improve performance in finding the shortest path between two points.

Since Dijkstra's proposal, many algorithms have been created being able to find the shortest path with the lowest computational cost.
A* \cite{grid}, Bi A* \cite{bi_a_start}, Breadth-first \cite{kurant2010bias}, Best-First \cite{dechter1985generalized} are a few of the search algorithms that exist for path planning. They have peculiarities that tackle different problems, and, therefore, are useful and important in many areas. These algorithms, however, become extremely costly when applied to large environments or environments with dynamic objects \cite{mohanan2018survey}. All these algorithms are detailed in the theoretical foundation section \ref{TF}.

The problem of the increased computational cost when increasing the amount of information is a problem that affects several fields of research, for example, pattern recognition \cite{pr}, computer vision \cite{cp}, text mining \cite{tm}. A very well-known approach to avoid this problem is to reduce dimensionality by discarding irrelevant information to the task.

Principal component analysis (PCA) is a mathematical procedure based on orthogonal transformation to convert data into a set of values of linearly unrelated variables called principal components. The number of principal components is always less than or equal to the number of original variables \cite{pca}.

Truncated Singular value decomposition (TSVD) This algorithm use means of TSVD to performs linear dimensionality reduction. Differently of PCA, this solution does not center the data before computing the singular value decomposition \cite{tsvd}.

Non-negative matrix factorization (NMF) Creates two non-negative matrices (W, H). The product of these matrices is an approximation of the non-negative input data. This method is used for dimensionality reduction \cite{nmf}.

These solutions significantly reduce dimensionality, keeping enough information to accomplish some tasks. However, a limitation of these approaches is since they reduce dimensionality using only linear correlation \cite{mpca}. Therefore, in \cite{NOIS}, the authors built a deep learning model able to reduce dimensionality using non-linearity correlation, named Convolutional Neuronal Network (CNN) Encoder. Their method removes mostly the useless information of the input data, including in dynamic environments. Eliminate useless information for path planning problem means to remove the paths which do not connect the start point and the goal point.

In this work, we perform an in-depth evaluation of the application of their proposed CNN Encoder to decrease the time spent by different path planning algorithms, showing the efficiency of the proposal to improve various path planning solutions.




\section{Material and Methods}

\subsection{Theoretical Foundation}
\label{TF}

\subsubsection{Dijkstra's algorithm}

The Dijkstra's algorithm proposed in 1956 by Edsger W. Dijkstra \cite{misa2010interview} is a path planning algorithm based on graph search. It solves the single-source shortest path problem for a graph with non-negative edge path costs, producing a shortest-path tree, it can be defined as follows:

\begin{algorithm}[h]
    \SetAlgoLined
    \KwResult{H, {dist} }
    initialization\;
    \For{each vertex $v$ in $V_G$}{
        $dist_{v} \gets \infty$\;
        $parent_{v} \gets NIL$\;
    }
    $dist_{s} \gets 0$\;
    \While{$Q \neq \emptyset$}{
        $u \gets ExtractMin(Q)$\;
        \For{each edge $e = (u,v)$}{
            $dist_{v} \gets \infty$\;
            \If{$dist_{v} > dist_{u} + weight_{e}$}{
                $dist_{v} \gets dist_{u} + weight_{e}$\;
                $parent_{v} \gets u$
            }
        }
    }
    $H \gets (V_G, \emptyset)$\;
    \For{each vertex $v in V_G,\ v \neq s$}{
    $E_H \gets E_H \cup \{(parent_{v}, v)\}$\;
    }
    \Return $H, dist$
\caption{Dijkstra's algorithm}
\end{algorithm}


\subsubsection{A *}
The A* algorithm \cite{hart1968formal} was proposed to solve the limitations of Dijkstra's algorithm, and to overcome it in the time spent to find the shortest path. This solution uses heuristics to be faster. It is defined as:

\begin{algorithm}
\SetAlgoLined
\KwResult{$Shortest\ path$}
$closedset\gets \emptyset$\;
$openset\gets start$\;
$came from\gets empty map$\;

$g score[start]\gets 0$\;

$f score[start]\gets g\ score[start] + heuristic \ cost\ estimate(start, goal)$\;

\While{$openset \neq \emptyset$}{
  $current \gets the\ node\ in\ openset\ having\ the\ lowest\ f score\ value$\;
\If{$current = goal$}{
\Return $Reconstruct\ Path(came\ from,goal)$\;
}
$remove\ current\ from\ openset$\;
$add\ current\ to\ closedset$\;

\For{$each\ neighbor\ in\ neighbor\ nodes(current)$}{
 \If{$neighbor\ in\ closedset$}{
    $continue$\;
}
$tentative\ g\ score\ \gets g\ score[current] + dist between(current,neighbor)$\;
}
\If{$neighbor\ not\ in\ openset\ or\ tentative\ g\ score\ <\ g\  score[neighbor]$}{
    $came from[neighbor] \gets current$\;
    $g score[neighbor] \gets tentative g score$\;
    $f score[neighbor] \gets g score[neighbor] + heuristic cost estimate(neighbor, goal)$\;
    \If{$neighbor\ not\ in\ openset$}{
        $add\ neighbor\ to\ openset$\;
    }
\Return $failure$
    
}
}

\caption{A* algorithm}
\end{algorithm}

\subsubsection{Bi A*}
Bidirectional A* search is a graph search algorithm that finds the shortest path from an initial vertex to a goal vertex in a directed graph running two simultaneous searches \cite{bi_a_start}. It can be described as: 

\begin{algorithm}
\SetAlgoLined
\KwResult{$H, dist$}
$A start1 \gets A start(start,goal)$\;
$ A start2 \gets A start(goal,start)$\;
\While{$A start1.current \neq A start2.current$}{
$path\ one \gets A start1.run()$\;
$path\ two \gets A start2.run()$\;
}
$shortest\ path \gets path\ one + path\ two$\;
$H \gets (shortest\ path)$\;
\Return $H$
\caption{Bi A*}
\end{algorithm}

\subsubsection{Breadth-first}
The Breadth-first Search \cite{kurant2010bias} is a classic graph search algorithm, and it works by expanding and systematically exploring a given node and progressively redoing the same procedure for all its neighbors. At each iteration, the last one explored, but not expanded, or visited, is selected. Also, this algorithm discovers all nodes that are a certain distance from the start.

\begin{algorithm}[h!]
\SetAlgoLined
\KwIn{$ nodes, source $}
\KwResult{$tree$}
$start\gets {\{source\}}$\;
$next\gets {\{\}}$\;
$parents\gets {[-1, -1, ..., -1]}$\;
 \While{$start \neq {\{\}}$}{
    \For{ $v \in start$}{
    \For{$n \in neighbors[v]$}{
        \If{$parents[n] = -1$}{
            $parents[v] \gets v$\;
            $next \gets next \cup \{n\}$\;
            }
        }
    }
    $start \gets {next}$\;
    $next \gets {\{\}}$\;
}

\Return $tree$
\caption{Breadth-first algorithm}
\end{algorithm}
\BlankLine
\BlankLine
\subsubsection{Best-First}
Based on the different strategies for solving search problems, the Best-First \cite{dechter1985generalized} algorithm is one of the most popular in the literature. Given the heuristic function \(F(n) = h(n)\), which is applied equally throughout the search space, the algorithm aims to use this to quantify the value of each candidate exploited during the process and, thus, continue the exploration until reach the point of interest.
\BlankLine
\BlankLine

\begin{algorithm}[h!]
\SetAlgoLined
\KwIn{$ Graph, start, goal $}

$PriorityQueue \gets {\{\}}$\;
$PriorityQueue \gets {start}$\;

\While{$PriorityQueue \neq {\{\}}$}{
    $u \gets removeMin(PriorityQueue)$\;

    \If{$u = goal $}{
        $exit$;  
    }

    \For{$each\ neighbor\ v\ of\ u$}{
         \If{$v\ is\ unvisited$}{
            $mark\ v\ as\ visited$\;
            $PriorityQueue \gets {v}$\;
        }
    }
    $mark\ u\ as\ exploited$\;

}

\caption{Best-First algorithm}
\end{algorithm}

\subsection{Path Planning Database}

Following the research of Janderson et al. previous work and thus demonstrating the efficiency of their proposal concerning the conventional approaches, we use the image database that was proposed by them, to expand previous results through the comparison with other solutions present in the literature. It contains various formats, distributed in five different scenarios. They have variations of start, goal points for each instance in the database; consequently, the possibilities of paths to be taken. In Figure \ref{database}, it is possible to see an instance of each scenario.

Using their database is possible to perform a more detailed analysis of the model's ability to generalize its responses, as well as its level of efficiency when compared to other solutions in different scenario configurations. As mentioned before, their database contains five scenarios, where each one has a total of 10000 scenes, which are RGB images with a resolution of 60 x 60 pixels. The variation between them is due to the random positioning of obstacles; which simulates physical obstacles. Also, for each one of the images, there is a label, being a GrayScale 60 x 60 image, which contains the shortest path of the scene.

\begin{figure}[htbp]
\centerline{\includegraphics[width=.4\textwidth]{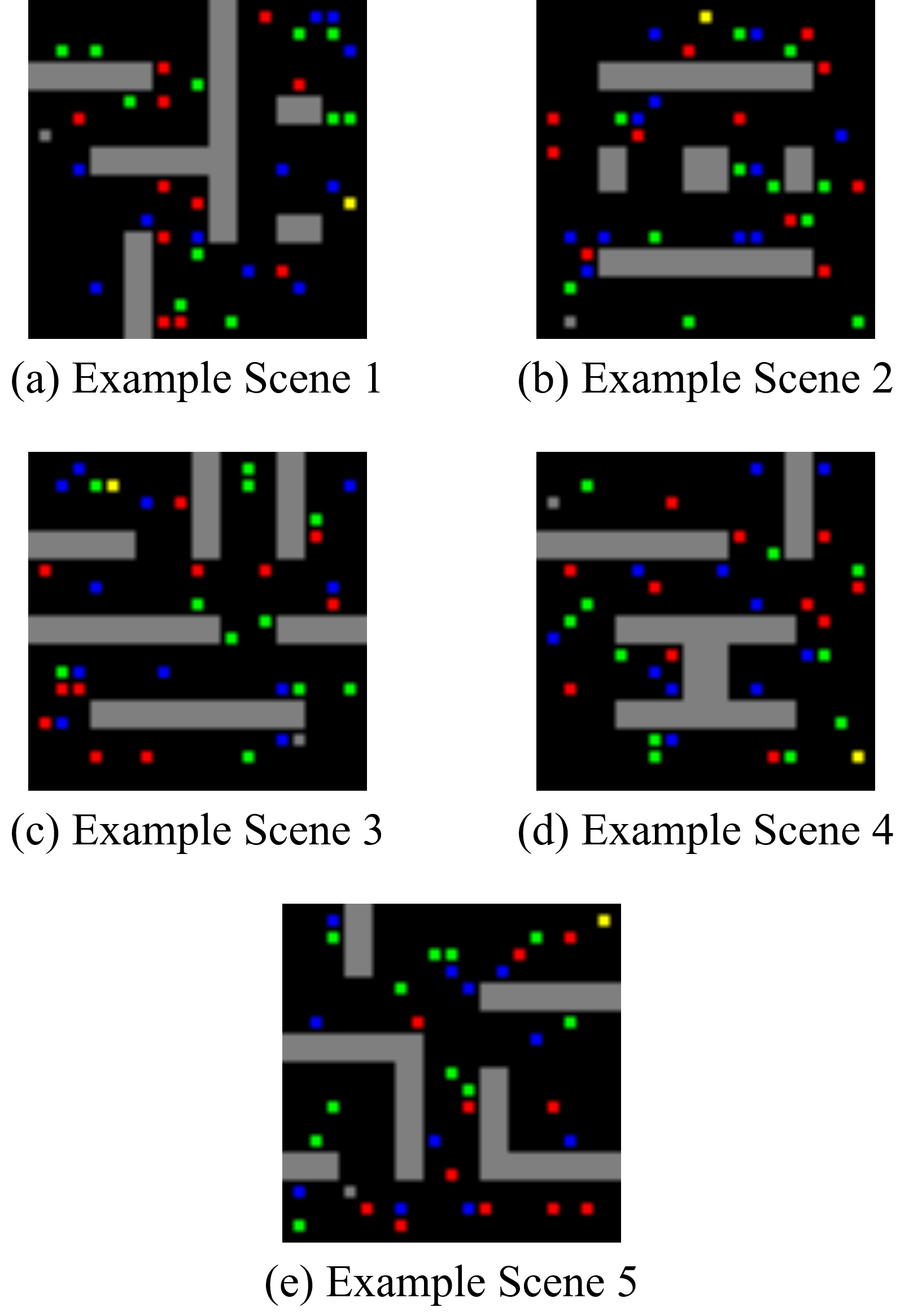}}
\caption{The instances of the used database. The yellow pixels represents the start and, the gray ones, the end of the path.}
\label{database}
\end{figure}

\subsection{The Convolutional Neural Networks Encoder}
Autoencoders are being used to code data information in unsupervised learning \cite{firstae}. They are trained to reconstruct the input data using fewer data than the original input; this way, many times, they can eliminate useless information. On the other hand, Convolutional Neural Networks has an excellent capability to extract high-level features in tasks of deep learning and computational vision problems\cite{lecun2015deep}. Trying to get the best of each model, Janderson et al. built a CNN Encoder to reduce the dimensionality of the data. This solution can eliminate useless routes from 2D maps \cite{NOIS}. 

\subsubsection{The model architecture}
The architecture used was obtained through a testing process, where its construction took place through adjustments based on the results generated. Finally, the architecture reached can be analyzed in Table \ref{arq}.

\begin{table}[h!]
\centering
\caption{Architecture of the CNN Encoder}
\label{arq}
\scalebox{0.8}{%
\begin{tabular}{|c|c|c|c|c|c|c|}
\hline
\multicolumn{2}{|c|}{Layer} & Filters & Kernel Size & Activation & Batch Norm & Dropout \\ \hline
1         & Image           & -       & -           & -          & -          & -       \\ \hline
2         & Conv            & 64      & 3x3         & ReLu       & True       & -       \\ \hline
3         & Conv            & 128     & 3x3         & ReLu       & False      & 30\%    \\ \hline
4         & Max-Pool        & -       & 3x3         & -          & -          & -       \\ \hline
5         & Conv            & 256     & 3x3         & ReLu       & True       & -       \\ \hline
6         & Conv            & 512     & 3x3         & ReLu       & False      & 30\%    \\ \hline
7         & Dense           & 256     & -           & LeakyReLU  & True       & 30\%    \\ \hline
8         & Dense           & 512     & -           & LeakyReLU  & True       & 30\%    \\ \hline
9         & Dense           & 1024    & -           & LeakyReLU  & True       & 30\%    \\ \hline
10        & Dense           & 3600    & -           & Tanh       & False      & 30\%    \\ \hline
\end{tabular}}
\end{table}

\subsection{Experimental Setup}
\label{expProcedure}

\begin{figure}[htbp]
\centerline{\includegraphics[width=.4\textwidth]{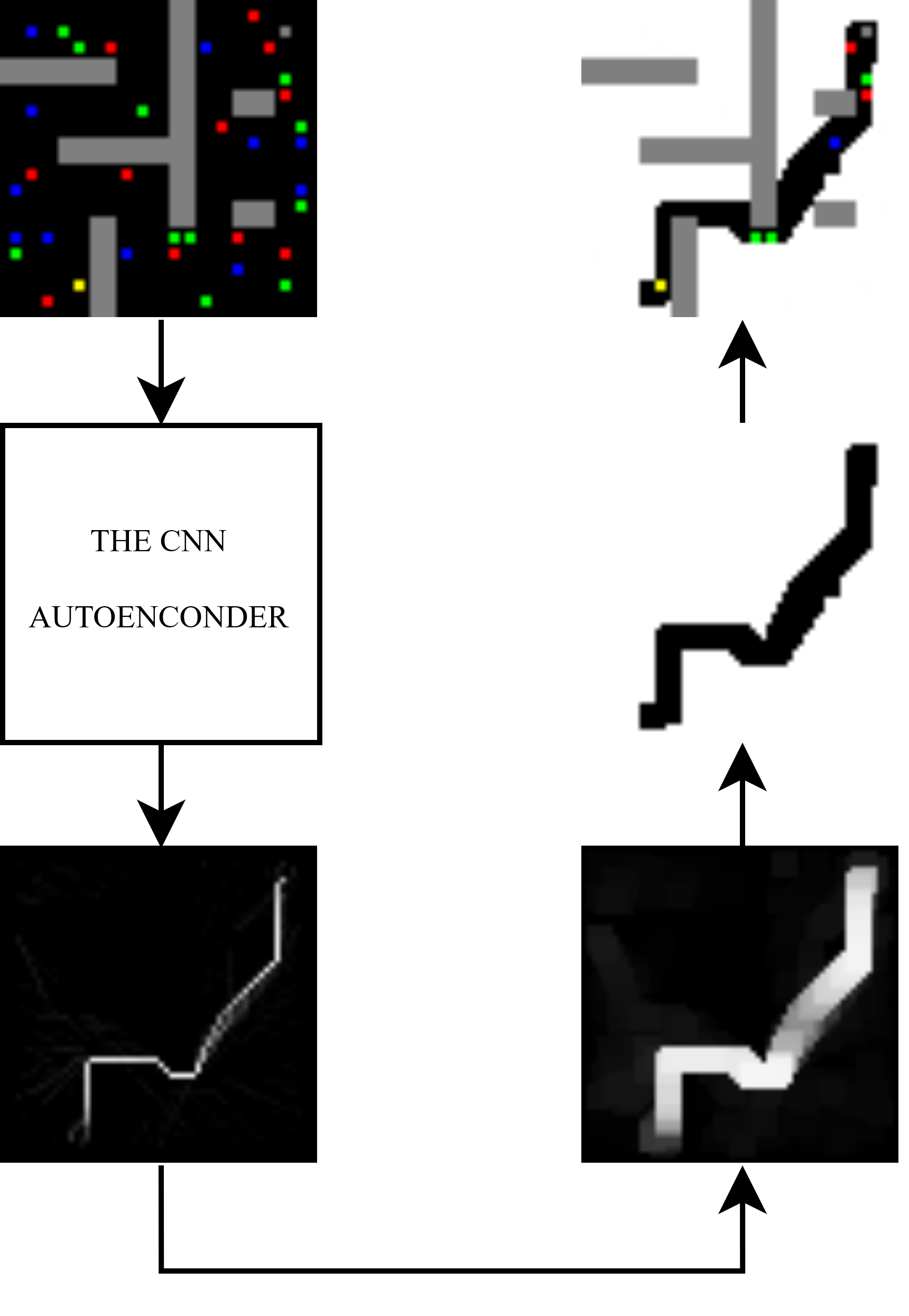}}
\caption{Representation of the processing applied to the output of CNN Encoder. Here it is possible to analyze that approach has achieved a considerable reduction in the search space. In both the first and the last stages of the process, the black pixels represent the walking paths for the search algorithm.}
\label{database}
\end{figure}

In this section, we describe the metrics used to evaluate the obtained results, how the database was manipulated, and the output processing. We also specify the hardware characteristics of the desktop computer used to perform the experiments; this is important because the hardware influences some experiments.

\subsubsection{Hardware Specification}

The algorithms were written in python 3.7 and implemented in an Intel i5-8400 six-core, with a base frequency of 2.80GHz and 8GB of RAM. More details are depicted in Table \ref{tab:processorSpecs}.

\begin{table}[htb]
\centering
\caption{Specifications of the desktop computer used to run the experiments.}
\label{tab:processorSpecs}
\begin{tabular}{lc}
\hline
Model &  \multicolumn{1}{l}{Intel(R) Core(TM) i5-8400GHz} \\
Number of Cores & 6 \\
Number of Threads & 6 \\
Base Frequency & 2.80GHz \\
Cache Size & 9MB \\
RAM & 8GB \\
\end{tabular}%
\end{table}	

\subsubsection{Metrics}

\begin{itemize}
\item Number of iterations.
Represents the number of attempts until the algorithm found the shortest path.
\item The path planning algorithm time.
Represents the time in seconds that the algorithm spent to find the shortest path.
\item CNN Encoder and preprocess output time.
Represents the time in seconds that the CNN Encoder spent to predict the data input plus the time to preprocess it.
\item Total time.
\end{itemize}

\subsubsection{Database Division}

Percentage Split

\begin{itemize}
    \item Train 80\%
    \item Validation 10\%
    \item Test 10\%
\end{itemize}

\section{Experimental results}
\label{secao_3}

To evaluate this work, we applied the path planning algorithms mentioned in the theoretical foundation to the path planning database, and compare the average number of iterations when the CNN Encoder is combined or do not with these algorithms. Also, we compare the average of the time spent with and without using the proposal. The results were obtained using the test set (1000 images for each scene).

To facilitate the visualization of the results, they were compiled and divided into three tables. Table \ref{results_iterations} shows the number of interactions of each algorithm for each scenario separately; this is important to analyze the model behaviour in different scenarios. Also, Table \ref{results_iterations} shows the percentage of improvement between the algorithms when using or not using the model. On the other hand, Table \ref{results_time} shows the spent time of the algorithms to find the shortest path for each scenario. Also, the improvement time with and without the model was calculated.

In the conventional A* algorithm, we obtained an average improvement of 59.87\% in terms of number of iterations. Looking at the execution time, with the application of the solution, we achieved an average improvement of 49.87\%.

Going to the Best-First search algorithm, we achieved an average improvement of 53.87\% in terms of number of iterations, also, in execution time, an average improvement of 16.06\%.

Using a variation of the first tested algorithm, Bi A*, an average improvement of 56.12\% was presented in terms of number of iterations, in the execution time, an average improvement of 40.88\%.

We also used the Breadth-first algorithm, which is very widespread in the literature, we managed to achieve an average improvement of 83.77\% in terms of number of iterations, in the execution time, an average improvement of 65.46\%.

Finally, we applied the proposal to the Dijkstra algorithm, we obtained an average improvement of 84.12\% in the number of iterations and looking at the execution time, there was an average improvement of 81.24\%.

\begin{figure}[htbp]
\centerline{\includegraphics[width=.4\textwidth]{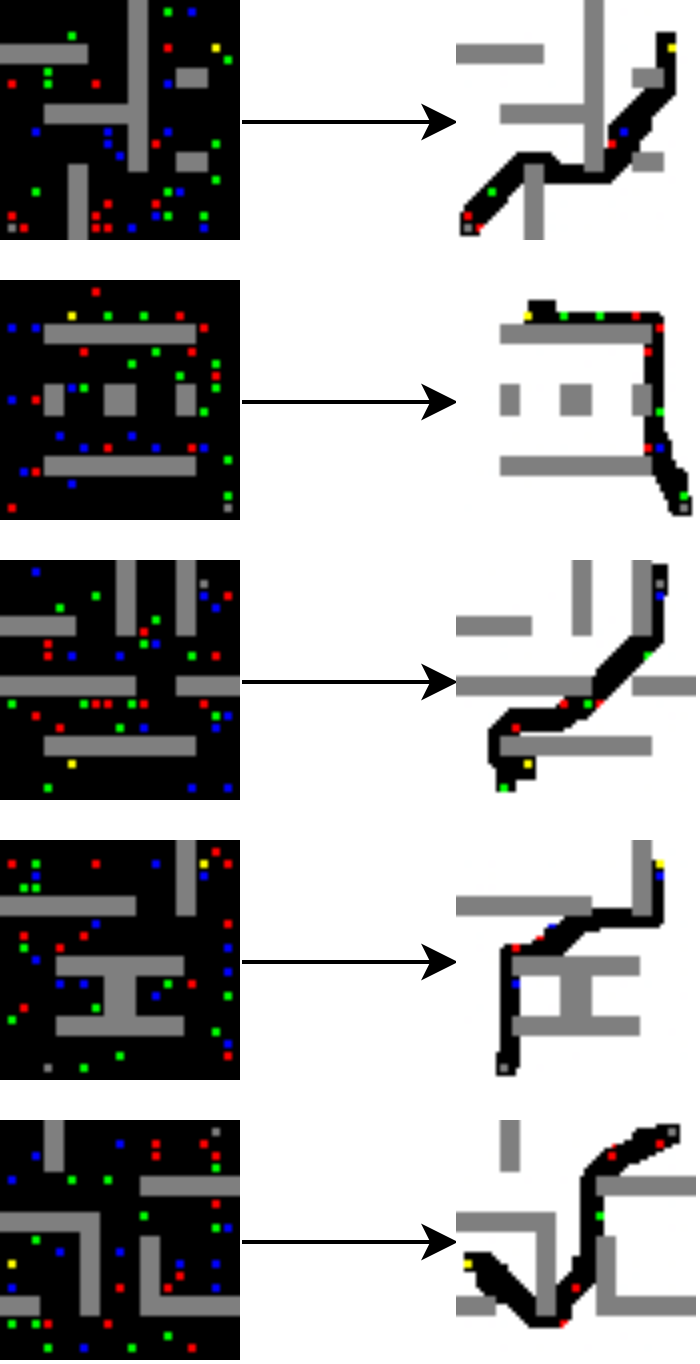}}
\caption{Comparison of the search space in an instance of each class, in the left column, the original scenes, in the right, the new search space after the application of the solution.}
\label{compare}
\end{figure}

In Table \ref{results_compiled}, it is possible to see a compilation of our results, containing a summary comparison between the standard execution of the search algorithms and their new results after the addition of our proposal. Also, it is possible to see some qualitative results in Figure \ref{compare}.

\begin{table}[!h]
\centering
\caption{Comparison between the number of iterations using only the path planning algorithms and the result after adding the CNN Encoder.}
\label{results_iterations}
\scalebox{0.8}{%
\begin{tabular}{|l|l|l|l|l|l|}
\hline
                            & Scene 1 & Scene 2 & Scene 3 & Scene 4 & Scene 5 \\ \hline
A*                          & 910.95  & 623.33  & 449.56  & 616.71  & 584.32  \\ \hline
CNN Encoder + A*            & 301.42  & 246.70  & 221.70  & 254.06  & 254.07  \\ \hline
Improvement                 & 66.91\% & 60.42\% & 50.68\% & 58.80\% & 56.52\% \\ \hline
\textbf{Total Improvement}  & \multicolumn{5}{c|}{\textbf{59.87\%}}           \\ \hline
Best-first                  & 263.07  & 86.01   & 119.24  & 162.42  & 174.1   \\ \hline
CNN Encoder + Best-first    & 93.45   & 75.64   & 63.15   & 67.69   & 71.34   \\ \hline
Improvement                 & 64.48\% & 12.05\% & 47.04\% & 58.32\% & 59.03\% \\ \hline
\textbf{Total Improvement}  & \multicolumn{5}{c|}{\textbf{53.87\%}}           \\ \hline
Bi A*                       & 946.31  & 491.44  & 458.24  & 665.34  & 499.88  \\ \hline
CNN Encoder + Bi A*         & 341.61  & 254.85  & 227.30  & 272.18  & 247.39  \\ \hline
Improvement                 & 63.90\% & 48.14\% & 50.40\% & 59.09\% & 50.51\% \\ \hline
\textbf{Total Improvement}  & \multicolumn{5}{c|}{\textbf{56.12\%}}           \\ \hline
Breadth-first               & 2268.59 & 2591.56 & 2188.41 & 2376.08 & 2286.77 \\ \hline
CNN Encoder + Breadth-first & 413.48  & 377.33  & 384.07  & 368.17  & 359.62  \\ \hline
Improvement                 & 81.77\% & 85.44\% & 82.45\% & 84.51\% & 84.27\% \\ \hline
\textbf{Total Improvement}  & \multicolumn{5}{c|}{\textbf{83.75\%}}           \\ \hline
Dijkstra                    & 2284.51 & 2669.13 & 2285.65 & 2405.46 & 2343.36 \\ \hline
CNN Encoder + Dijkstra      & 413.82  & 377.70  & 384.33  & 368.38  & 359.83  \\ \hline
Improvement                 & 81.89\% & 85.85\% & 83.18\% & 84.69\% & 84.64\% \\ \hline
\textbf{Total Improvement}  & \multicolumn{5}{c|}{\textbf{84.12\%}}           \\ \hline
\end{tabular}}
\end{table}

\begin{table}[!h]
\centering
\caption{Comparison between the spent time using only the path planning  and the result after adding the CNN Encoder.}
\label{results_time}
\scalebox{0.8}{%
\begin{tabular}{|l|l|l|l|l|l|}
\hline
                            & Scene 1 & Scene 2  & Scene 3 & Scene 4 & Scene 5 \\ \hline
A*                          & 0.027   & 0.021    & 0.013   & 0.018   & 0.017   \\ \hline
CNN Encoder + A*            & 0.010   & 0.010    & 0.009   & 0.009   & 0.009   \\ \hline
Improvement                 & 61.25\% & 51.61\%  & 28.79\% & 49.60\% & 46.13\% \\ \hline
\textbf{Total Improvement}  & \multicolumn{5}{c|}{\textbf{49.87\%}}            \\ \hline
Best-first                  & 0.009   & 0.003    & 0.004   & 0.005   & 0.005   \\ \hline
CNN Encoder + Best-first    & 0.005   & 0.004    & 0.004   & 0.004   & 0.004   \\ \hline
Improvement                 & 34.94\% & -50.54\% & -1.42\% & 26.42\% & 24.96\% \\ \hline
\textbf{Total Improvement}  & \multicolumn{5}{c|}{\textbf{16.06\%}}            \\ \hline
Bi A*                       & 0.024   & 0.013    & 0.011   & 0.017   & 0.012   \\ \hline
CNN Encoder + Bi A*         & 0.010   & 0.009    & 0.008   & 0.009   & 0.008   \\ \hline
Improvement                 & 55.08\% & 30.37\%  & 26.83\% & 45.76\% & 30.87\% \\ \hline
\textbf{Total Improvement}  & \multicolumn{5}{c|}{\textbf{40.88\%}}            \\ \hline
Breadth-first               & 0.013   & 0.015    & 0.013   & 0.014   & 0.013   \\ \hline
CNN Encoder + Breadth-first & 0.005   & 0.004    & 0.004   & 0.004   & 0.004   \\ \hline
Improvement                 & 60.81\% & 69.08\%  & 62.42\% & 67.51\% & 66.69\% \\ \hline
\textbf{Total Improvement}  & \multicolumn{5}{c|}{\textbf{65.46\%}}            \\ \hline
Dijkstra                    & 0.043   & 0.056    & 0.045   & 0.049   & 0.048   \\ \hline
CNN Encoder + Dijkstra      & 0.009   & 0.009    & 0.009   & 0.008   & 0.008   \\ \hline
Improvement                 & 77.07\% & 83.73\%  & 79.57\% & 82.33\% & 82.54\% \\ \hline
\textbf{Total Improvement}  & \multicolumn{5}{c|}{\textbf{81.24\%}}            \\ \hline
\end{tabular}}
\end{table}

\begin{table}[!h]
\centering
\caption{The average of time and interaction of all scenarios for each model}
\label{results_compiled}
\scalebox{1}{%
\begin{tabular}{|l|l|l|}
\hline
                                      & Iterations     & Time (s)              \\ \hline
A*                                    & 636.97          & 0.0194748744          \\ \hline
\textbf{CNN Enconder + A*}            & \textbf{255.59} & \textbf{0.0097622888} \\ \hline
Best-first                            & 160.98          & 0.0056170788          \\ \hline
\textbf{CNN Enconder + Best-first}    & \textbf{74.25}  & \textbf{0.00471486}   \\ \hline
Bi A*                                 & 612.24          & 0.0158272148          \\ \hline
\textbf{CNN Enconder + Bi A*}         & \textbf{268.67} & \textbf{0.0093569152} \\ \hline
Breadth-first                         & 2342.28         & 0.0141633262          \\ \hline
\textbf{CNN Enconder + Breadth first} & \textbf{380.53} & \textbf{0.0048917796} \\ \hline
Dijkstra                              & 2397.62         & 0.0486876686          \\ \hline
\textbf{CNN Enconder + Dijkstra}      & \textbf{380.81} & \textbf{0.0091321876} \\ \hline
\end{tabular}}
\end{table}

\section{Conclusion}
\label{secao_5}

This work aimed to show that it is possible to improve the performance of path planning algorithms using a CNN Encoder to eliminate useless routes.

From the results obtained, we can assume that it is more advantageous to apply the CNN encoder to the existing path planning techniques. That is, the proposal was able to reduce the time to find the shortest path with all analyzed algorithms. In fact that CNN Encoder can eliminate routes in scenarios with fixed and dynamic obstacles, which may help in research with robotic navigation. As such, our contribution is to validate the architecture's efficiency with different solutions from the path planning literature.

As future work, we hope to check the proposed model for the creation of a socially aware motion planning algorithm \cite{diff_soci}. Also, we intend to combine new Deep Learning features with improving the architecture, which may reduce the response time even further.



\begin{footnotesize}

\bibliographystyle{unsrt}
\bibliography{references}

\end{footnotesize}

\vspace{12pt}

\end{document}